\documentclass[preprint]{article}
\usepackage{spconf}
\usepackage{amsmath,graphicx,hyperref,bm,nth,xspace,subcaption,stfloats,amssymb}
\usepackage[most]{tcolorbox}

\newcommand{\directname}{\textsc{Direct}}
\newcommand{\cotname}{\textsc{CoT}}

\newcommand{\directbaseline}{\directname\textsubscript{\textsc{Base}}\xspace}
\newcommand{\cotbaseline}{\cotname\textsubscript{\textsc{Base}}\xspace}

\newcommand{\directaugmentedhundred}{\directname\textsubscript{\textsc{aug100}}\xspace}
\newcommand{\directaugmentedxx}{\directname\textsubscript{\textsc{augXX}}\xspace}
\newcommand{\directaugmented}{\directname\textsubscript{\textsc{aug}}\xspace}

\newcommand{\cotaugmented}{\cotname\textsubscript{\textsc{aug}}\xspace}
\newcommand{\cotpaugmented}{\cotname\textsuperscript{\textdagger}\textsubscript{\textsc{aug}}\xspace}
\newcommand{\cotaugmentedtwenty}{\cotname\textsubscript{\textsc{aug20}}\xspace}

\newcommand{\cotaugmentedxx}{\cotname\textsubscript{\textsc{augXX}}\xspace}
\newcommand{\cotpaugmentedxx}{\cotname\textsuperscript{\textdagger}\textsubscript{\textsc{augXX}}\xspace}

\title{Revisiting Direct Speech-to-Text Translation with Speech LLMs: Better Scaling than CoT Prompting?}
\name{Oriol Pareras$^{1}$, Gerard I. Gállego$^{1,2}$, Federico Costa$^{1, 2}$, Cristina España-Bonet$^{1,3}$, Javier Hernando$^{1,2}$}
\address{
$^1$Barcelona Supercomputing Center, Spain
$^2$Universitat Politècnica de Catalunya, Spain\\
$^{3}$DFKI GmbH, Saarland Informatics Campus, Saarbrücken, Germany \\
\texttt{oriol.pareras@bsc.es, gerard.gallego@bsc.es}
}

\renewenvironment{thebibliography}[1]{%
  \begin{oldthebibliography}{#1}%
  \setlength{\parskip}{0pt}%
  \setlength{\itemsep}{0pt plus 0.3ex}%
}{%
  \end{oldthebibliography}%
}

\begin{document}
\copyrightnotice{
    \begin{minipage}{\textwidth}
        \centering
        \textcolor[gray]{0.5}{\footnotesize{\copyright 2026 IEEE. Personal use of this material is permitted. Permission from IEEE must be obtained for all other uses, in any current or future media, including reprinting/republishing this material for advertising or promotional purposes, creating new collective works, for resale or redistribution to servers or lists, or reuse of any copyrighted component of this work in other works. To appear in {\it Proc.\ ICASSP 2026, May 04-08, 2026, Barcelona, Spain}
        }}
    \end{minipage}
}

%
\maketitle
\begin{abstract}
Recent work on Speech-to-Text Translation (S2TT) has focused on LLM-based models, introducing the increasingly adopted Chain-of-Thought (CoT) prompting, where the model is guided to first transcribe the speech and then translate it. CoT typically outperforms direct prompting primarily because it can exploit abundant Automatic Speech Recognition (ASR) and Text-to-Text Translation (T2TT) datasets to explicitly model its steps. In this paper, we systematically compare CoT and Direct prompting under increasing amounts of S2TT data. To this end, we pseudo-label an ASR corpus by translating its transcriptions into six European languages, and train LLM-based S2TT systems with both prompting strategies at different data scales. Our results show that Direct improves more consistently as the amount of data increases, suggesting that it may become a more effective approach as larger S2TT resources are created.
\end{abstract}
\section{Introduction}
\label{sec:intro}

End-to-End (E2E) S2TT is increasingly preferred over the traditional cascaded pipeline of ASR followed by T2TT. Unlike cascaded approaches, E2E systems avoid error propagation across modules and can in principle exploit acoustic and prosodic cues that are lost in intermediate transcripts. Recent studies show that the performance of E2E systems is approaching that of cascaded methods, with some works even reporting improvements beyond them~\cite{du-etal-2025-making}.

A popular strategy for building E2E systems is to adapt a backbone LLM already fine-tuned for T2TT to the speech modality~\cite{wu-et-al-ondecoderonly,chen-etal-2024-llast}. Within this framework, recent works adopt \cotname\xspace prompting, where the model is guided to first transcribe and then translate~\cite{hu2025chain,huang_speech_2023,du_cot-st_2024}. \cotname\xspace has gained attention because it outperforms direct translation (without intermediate steps). More importantly, \cotname\xspace allows to leverage widely available ASR and T2TT data by explicitly modeling these steps.

In this work, we revisit this comparison and argue that the advantage of \cotname\xspace is mainly tied to the large quantity of ASR and T2TT data currently used for training. We show that as S2TT data becomes increasingly available, direct translation may emerge as a stronger alternative. To the best of our knowledge, this is the only systematic study that compares \directname\xspace and \cotname\xspace prompting under large-scale pseudo-labeled S2TT data with this much language coverage.

\begin{figure}[!t]
  \centering
  \includegraphics[width=0.9\linewidth]{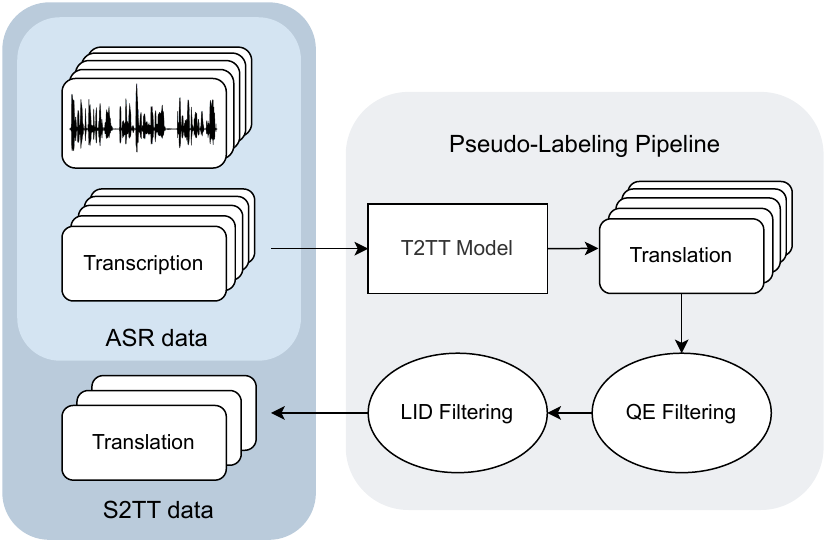}
  \vspace{-3pt}
  \caption{Overview of the pseudo-labeling pipeline.}
  \label{fig:pipeline}
  \vspace{-10pt}
\end{figure}

The dominant trend for creating S2TT resources is to translate existing ASR corpora into target languages~\cite{pino-etal-2019-harnessing,wang20ia_interspeech}. Following this approach, we construct a pseudo-labeled dataset using a translation pipeline (Figure~\ref{fig:pipeline}), and train models with both \directname\xspace and \cotname\xspace while scaling the amount of data. Our results demonstrate that \directname\xspace exhibits more stable scaling than \cotname\xspace.

This finding mirrors the evolution of ASR itself, where early phoneme- or grapheme-based pipelines were replaced by E2E models that directly map speech to text~\cite{hannun2014deep}. We presume that, by enforcing an intermediate transcription, \cotname\xspace may constrain the model to a fixed alignment between speech and text, which may limit the potential of E2E models of leveraging paralinguistic information. While this work does not explore such signals here, it highlights \directname\xspace prompting as a viable strategy and points toward future scenarios where richer speech annotations could further enhance E2E S2TT.

\section{Methodology}
\label{sec:method}

We systematically compare how \cotname\xspace and \directname\xspace prompting strategies scale in S2TT. To do so, we generate pseudo-labeled S2TT data (S2TT\textsubscript{pl}) and evaluate a series of models trained with varying amounts of it.

\subsection{Translation pseudo-labeling}
\label{subsec:pseudo-labeling}

We generate pseudo-labeled data by translating the transcriptions of an ASR dataset (Figure~\ref{fig:pipeline}). We use the same translation LLM employed as backbone for the S2TT model (see \S\nobreak\ref{subsec:stmodel}). To filter out generation errors or low-quality translations, we first score the samples with a Quality Estimation (QE) system. We use BLASER 2.0~\cite{dale-costa-jussa-2024-blaser} for this task because it provides reference-free, sentence-level translation scores that correlate strongly with human judgments. We compute the scores by evaluating the similarity of the transcription and translation with \texttt{blaser\_2\_0\_qe}\footnote{\scriptsize\url{https://huggingface.co/facebook/blaser-2.0-qe}} and remove samples below $3.75$, which was determined through inspection of samples across multiple languages.
Although BLASER 2.0 estimates translation quality, it is based on SONAR~\cite{Duquenne:2023:sonar_arxiv} embeddings, which encode speech and text from multiple languages into a shared space. Thus, it does not explicitly enforce the intended target language. We therefore apply a Language Identification (LID) system, using GlotLID v3~\cite{kargaran2023glotlid}, to compute language probabilities $\bm{p}=(p_{1},...,p_L)$ and discard samples with expected language probability $p_e < 0.5$.

\subsection{Speech Translation Model}
\label{subsec:stmodel}

We build an LLM-based E2E S2TT system for our experimentation, following a similar approach as~\cite{zhang_speechgpt_2023,rubenstein_audiopalm_2023}. Each speech utterance \( x \) is first encoded with a self-supervised model \( f_{\text{enc}} \), extracting continuous representations. These features are quantized into a sequence of discrete tokens:
\(
\bm{s} = (s_1, \ldots, s_T) = q\bigl(f_{\text{enc}}(x)\bigr),\quad s_t \in V_s
\),
where \( q(\cdot) \) is the quantization function and \( V_s \) is the speech token vocabulary. To adapt the pretrained LLM for speech inputs, its original vocabulary $V_o$ is expanded to $V = V_o \cup V_s$, alongside its corresponding token embedding matrix $E = \bigl[E_o;\,E_s\bigr] \in \mathbb{R}^{(|V|)\times d}$, where $E_s$ is randomly initialized.

Training is performed in two stages. In stage~1, the backbone LLM is frozen and only the token embedding layer is trained on speech utterances from the ASR corpora, using next-token prediction. In stage~2, the entire LLM is trained on the data described in \S\nobreak\ref{subsec:data}.

For the backbone model we use \texttt{salamandraTA\allowbreak-7B\allowbreak-Instruct},\footnote{\scriptsize \url{https://huggingface.co/BSC-LT/salamandraTA-7b-instruct}} a highly multilingual LLM fine-tuned for translation and proficient in 35 European languages. The speech encoder is mHuBERT\footnote{\scriptsize\url{https://huggingface.co/slprl/mhubert-base-25hz}} from TWIST~\cite{hassid_textually_2023}, together with its trained k-means quantizer. This encoder was chosen because: (1) it supports multilingual speech and (2) it downsamples to 25 Hz, halving the temporal resolution compared to standard HuBERTs~\cite{hsu_hubert_2021}. Discrete speech tokens are obtained by clustering the 11\textsuperscript{th}--layer representations into 500 groups.



\begin{figure}[t]
    \centering
    \small
    \begin{minipage}[t]{0.49\linewidth}
        \begin{tcolorbox}[
            colback=gray!8,
            colframe=gray!127,
            boxrule=0.6pt,
            title={CoT Prompt Template},
            coltitle=white,
            left=4pt,
            right=4pt,
            colbacktitle=gray!127,
            bottom=-6pt,
        ]
        \{audio\}\\
        Transcribe in \{src language\}\\
        \textbf{\{transcription\}}\\
        Translate to \{tgt language\}\\
        \textbf{\{translation\}}\\
        \end{tcolorbox}
    \end{minipage}    
    \begin{minipage}[t]{0.49\linewidth}
        \begin{tcolorbox}[
            colback=gray!8,
            colframe=gray!127,
            boxrule=0.6pt,
            title={Direct Prompt Template},
            coltitle=white,
            left=4pt,
            right=4pt,
            colbacktitle=gray!127,
            bottom=-6pt,
        ]
        \{audio\}\\
        Translate to \{tgt language\}\\
        \textbf{\{translation\}}\\
        \vspace{22pt}
        \end{tcolorbox}
    \end{minipage}
    \caption{Prompt templates (model outputs are in bold)}
    \label{fig:prompt}
    \vspace{-10pt}
\end{figure}

We use the prompts defined in~Figure~\ref{fig:prompt} for S2TT. We derive the templates from \cotname\xspace for ASR and T2TT, which allows better integration of model capabilities from these tasks into S2TT. The same template is applied during inference, and generation is performed with beam search using five beams.

\section{Experimental Setup}
\label{sec:experminental_setup}

\subsection{Data}
\label{subsec:data}

Our training framework includes four types of datasets: ASR, T2TT, S2TT and pseudo-labeled S2TT (S2TT\textsubscript{pl}) in six languages: Catalan (\textit{ca}), German (\textit{de}), English (\textit{en}), Spanish (\textit{es}), French (\textit{fr}), and Italian (\textit{it}). Table~\ref{tab:data-stats} reports the amount of data per language. For ASR, we use the train splits of Common Voice 21.0~\cite{ardila_common_2020} and Multilingual LibriSpeech~\cite{pratap2020MLSAL}, totaling approximately 6,000 and 48,900 hours, respectively. For T2TT, we use Wikimedia~\cite{TIEDEMANN12.463} samples with 5 to 100 words, filtered using the same pipeline applied to pseudo-labeled S2TT data. For S2TT, we consider the train splits of Europarl-ST v1.1~\cite{jairsan2020a} and CoVoST 2~\cite{wang_covost_2021}, totaling about 630 and 1,600 hours, respectively. Finally, S2TT\textsubscript{pl} data is generated by translating all Common Voice 21.0 samples into all possible target languages. We evaluate our models in CoVoST 2 and Fleurs~\cite{conneau_fleurs_2023} test sets.

\begin{table}[!h]
\centering
\small
\setlength{\tabcolsep}{3pt}
\renewcommand{\arraystretch}{1.2}
\caption{Amount of data per language and total target tokens (considering transcriptions and translations as target in S2TT). The number of S2TT\textsubscript{pl} hours reflects repeated utterances, as each utterance is paired with five target languages.}
\renewcommand{\arraystretch}{1.2}
\begin{tabular}{lccccccc}
\hline
 & ca & de & en & es & fr & it & tgt tok. \\
\hline
\multicolumn{8}{l}{\textit{Hours per source language}} \\
ASR       & 1,775 & 2,900 & 46,422 & 1,412 & 1,895 & 493 & 610M \\
S2TT      & 136 & 265 & 1,163 & 173 & 355 & 141 & 48M \\
S2TT\textsubscript{pl} & 8,875 & 4,670 & 8,810 & 2,470 & 4,090 & 1,230 & 384M \\
\hline
\multicolumn{8}{l}{\textit{Samples per target language}} \\
T2TT      & 289k & 48k & 76k & 668k & 508k & 160k & 134M \\
S2TT      & 289k & 341k & 585k & 54k & 50k & 48k & 48M \\
S2TT\textsubscript{pl}      & 1.35M & 2.05M & 2.63M & 2.18M & 2.10M & 2.36M & 384M \\
\hline 
\end{tabular}
\label{tab:data-stats}
\end{table}


\subsection{Experiments}
\label{subsec:experminents}

We train two baseline models using all the ASR, T2TT and S2TT data described in \S\nobreak\ref{subsec:data}, with direct (\textbf{\directbaseline}) and CoT (\textbf{\cotbaseline}) prompting.

To analyze the impact of scaling S2TT data, we incrementally add S2TT\textsubscript{pl} samples to each baseline, training 5 new systems with 20\%, 40\%, 60\%, 80\% and 100\% of the pseudo-labeled dataset. We refer to these models as \textbf{\directaugmentedxx} and \textbf{\cotaugmentedxx}, where \textsc{xx} is the percentage of pseudo-labeled data used. These models are trained independently, they are not checkpoints of a single experiment.

In the \cotname\xspace setting, each speech–transcription pair in the pseudo-labeled data is repeated six times, one for each target language. We hypothesize this repetition may harm the ASR performance, which is particularly critical for \cotname\xspace prompting. To validate this, we train two variants of \cotaugmentedxx: in \textbf{\cotpaugmentedxx}, the transcription step contributes to the loss, whereas in \textbf{\cotaugmentedxx} it does not.

We evaluate the models using Word Error Rate (WER) for ASR, and with BLEU~\cite{papineni_bleu_2002}, computed with SacreBLEU~\cite{post_call_2018},%
\footnote{signature:~\texttt{nrefs:1|case:mixed|eff:no|tok:13a|smooth:\allowbreak~exp|version:2.5.1}} as well as x\textsc{comet}~\cite{guerreiro-etal-2024-xcomet}\footnote{\scriptsize\url{https://huggingface.co/Unbabel/XCOMET-XL}} for T2TT and S2TT.

\begin{figure*}[!ht]
  \centering
  \begin{subfigure}{0.32\textwidth}
    \centering
    \includegraphics[width=\linewidth]{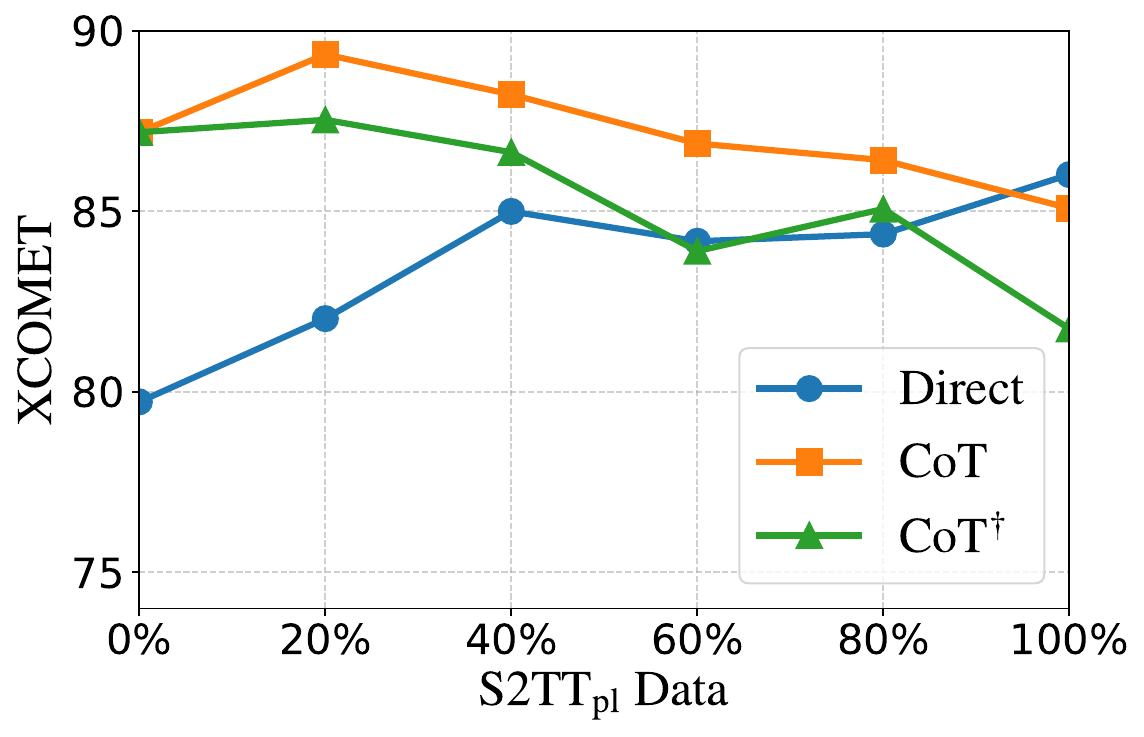}
    \caption{S2TT}
    \label{fig:s2tt_plot}
  \end{subfigure}
  \hfill
  \begin{subfigure}{0.32\textwidth}
    \centering
    \includegraphics[width=\linewidth]{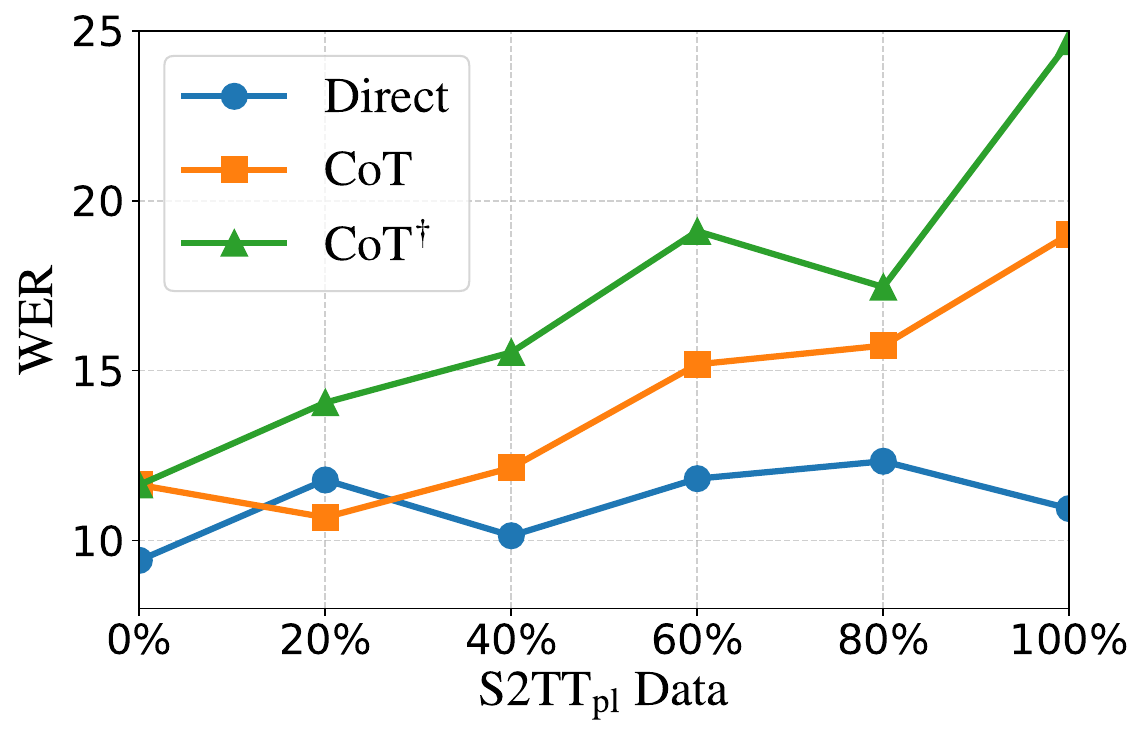}
    \caption{ASR}
    \label{fig:asr_plot}
  \end{subfigure}
  \hfill
  \begin{subfigure}{0.32\textwidth}
    \centering
    \includegraphics[width=\linewidth]{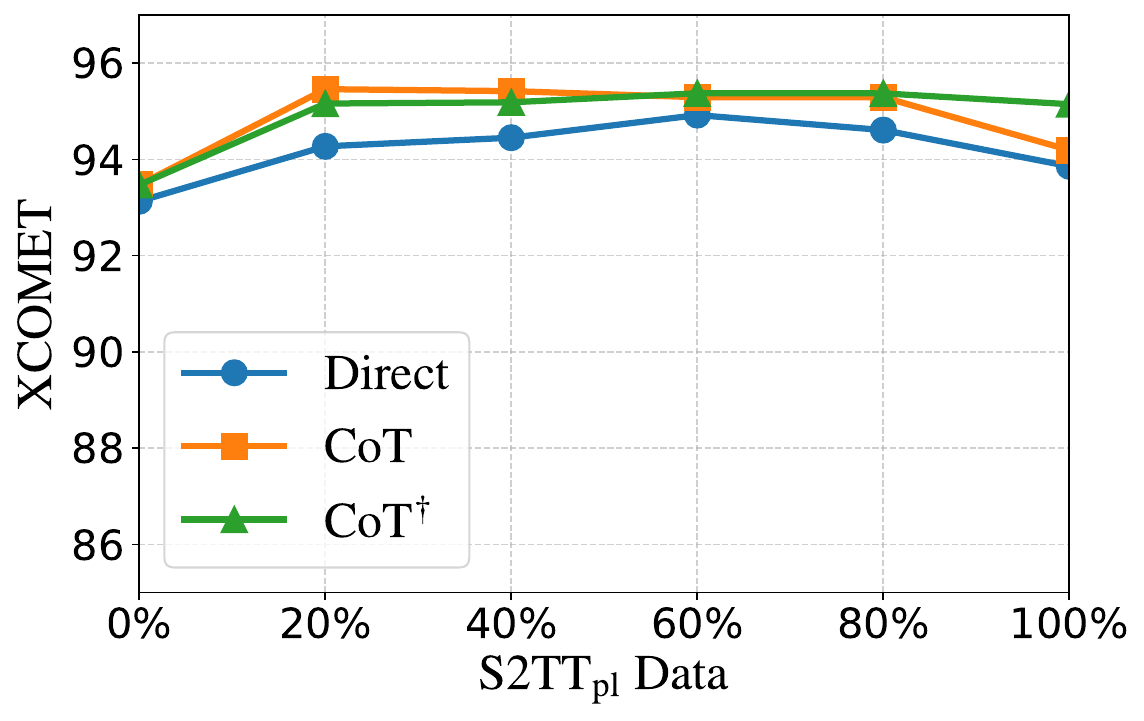}
    \caption{T2TT}
    \label{fig:t2tt_plot}
  \end{subfigure}
  \vspace{-3pt}
  \caption{Performance on the Fleurs benchmark when scaling the amount of S2TT\textsubscript{pl} data. S2TT and T2TT are evaluated with x\textsc{comet} ($\uparrow$) and ASR with WER ($\downarrow$). Note that $0\%$ corresponds to \textsc{base} variants and the rest to \textsc{aug\textsubscript{XX}}. Results are averaged across all the considered language directions.}
  \label{fig:x2x_plots}
  \vspace{-7pt}
\end{figure*}

\subsection{Training Details}

Both stage~1 and stage~2 (see \S\nobreak\ref{subsec:stmodel}) last one epoch and employ AdamW optimizer, a cosine learning rate scheduler, and gradient clipping with a maximum norm of~$1.0$.

In stage~1, we use a learning rate of $7 \cdot 10^{-5}$, a warm up during the first $3\%$ of updates, and set a maximum sequence length of~1024. Tokens from different sequences are packed into full-length sequences to improve efficiency, following~\cite{zhang_speechgpt_2023}. Training is run on 16~GPUs with a per-device batch size of~16, yielding an effective global batch size of~256.

In stage~2, we increase the maximum sequence length to~2048 (without sequence packing) and lower the learning rate to $4 \cdot 10^{-5}$, with warmup over the first $10\%$ of updates. We train on 16~GPUs with a per-device batch size of~32, resulting in an effective batch size of~512.

All experiments are implemented in the \texttt{transformers} library with DeepSpeed for distributed training, and run on NVIDIA H100 GPUs. To optimize memory usage, we apply mixed precision (bfloat16), gradient checkpointing, and Liger Kernel, which further improves efficiency and throughput.

\section{Results}
\label{sec:results}

\noindent\textbf{Initial \cotbaseline superiority}\;~We first verify that our baselines reproduce trends reported in prior work~\cite{hu2025chain}. As expected, Table~\ref{tab:fleurs-covost} shows that the \cotbaseline outperforms the \directbaseline across all languages, confirming the benefit of decomposing the task into transcription and translation in this scenario. The average baseline gap is approximately 5 BLEU points and 7 x\textsc{comet} points.

\begin{table}[h!]
\centering
\small
\setlength{\tabcolsep}{4pt} 
\renewcommand{\arraystretch}{1.1}
\caption{Baselines evaluation on FLEURS.}
\begin{tabular}{lccccccc}
\hline
 & \multicolumn{3}{c}{BLEU} & \multicolumn{3}{c}{x\textsc{comet}} \\
\cline{2-4} \cline{5-7}
 & x$\to$x & x$\to$en & en$\to$x & x$\to$x & x$\to$en & en$\to$x \\
\hline
\directbaseline & 21.04 & 22.80 & 30.32 & 80.6 & 79.7 & 86.0 \\
\cotbaseline    & \textbf{26.39} & \textbf{29.76} & \textbf{33.24} & \textbf{88.0} & \textbf{87.2} & \textbf{88.6} \\
\hline
\end{tabular}
\label{tab:fleurs-covost}
\end{table}

\noindent\textbf{\directname\xspace scales better than \cotname}\;~Figure~\ref{fig:s2tt_plot} shows that the two \cotaugmented variants peak at 20\% of the S2TT\textsubscript{pl} data and then degrade as more data is added. In contrast, \directaugmented consistently improves, achieving a better score with \directaugmentedhundred. Although the peaks of \cotpaugmented and \cotaugmented are still higher, \directaugmented demonstrates a steadier scaling trend, suggesting that this gap could be closed with additional data. These results indicate a potential advantage of \directname\xspace prompting when sufficient S2TT\textsubscript{pl} data is available.

\noindent\textbf{\cotname\xspace degrades ASR as data scales}\;~Figure~\ref{fig:asr_plot} shows that WER remains stable with \directaugmented across all S2TT\textsubscript{pl} data scales, whereas both \cotaugmented and \cotpaugmented gradually increase it by up to $+7.4$\% and $+13$\% relative to their baseline. These trends correlate with the steady decrease of S2TT performance, reflecting their critical reliance on the ASR sub-task. This is also observed at $20$\% of S2TT\textsubscript{pl} data, where \cotaugmented reaches its minimum WER alongside its highest x\textsc{comet} score. Additionally, \cotbaseline shows a 2-point WER reduction compared to its counterpart strategy. This confirms the findings of~\cite{wang20ia_interspeech}, indicating that direct S2TT data improve ASR performance.

Figure~\ref{fig:t2tt_plot} shows that \directname\xspace performs slightly worse on T2TT, likely due to the lower proportion of this task in the training recipe (in \directaugmented, S2TT\textsubscript{pl} samples do not contain the T2TT sub-task). Nevertheless, performance for all three methods remains stable around +2 x\textsc{comet} points. These results make \directname\xspace appealing, showing that ASR and T2TT remain stable as pseudo-labeled S2TT data scales.

\begin{figure*}[!ht]
  \centering
  \begin{subfigure}{0.32\textwidth}
    \centering
    \includegraphics[width=\linewidth]{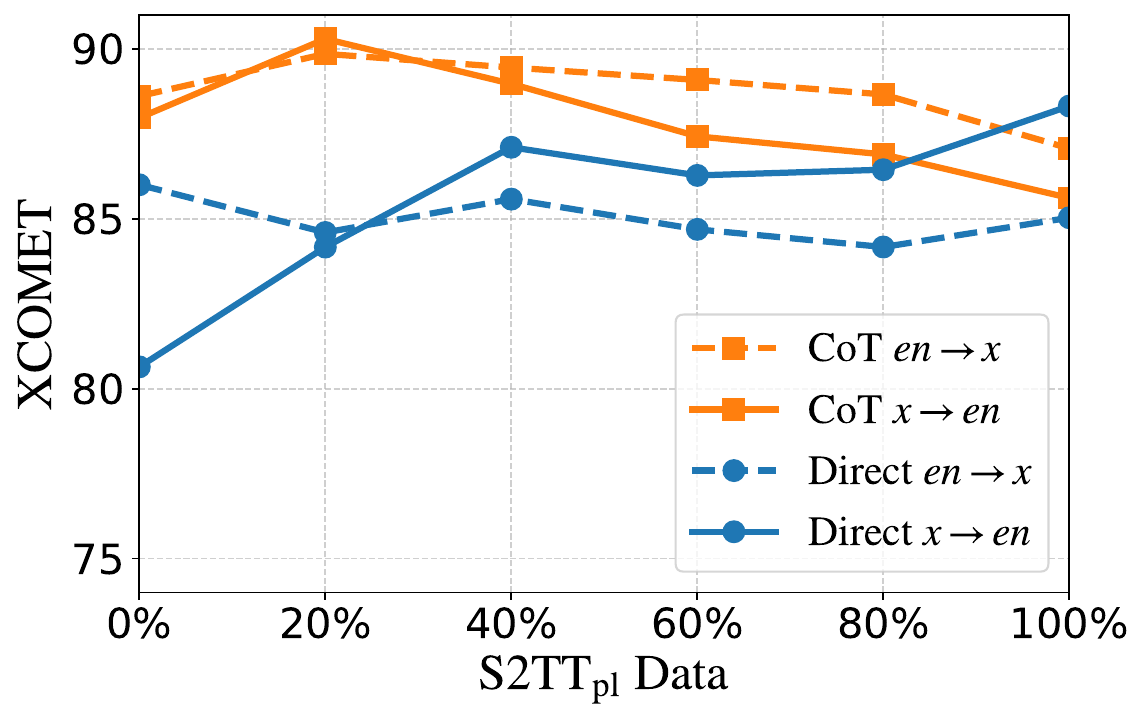}
    \caption{\textit{en$\rightarrow$x} and \textit{x$\rightarrow$en}}
    \label{fig:s2tt_plots_en}
  \end{subfigure}
  \hfill
  \begin{subfigure}{0.32\textwidth}
    \centering
    \includegraphics[width=\linewidth]{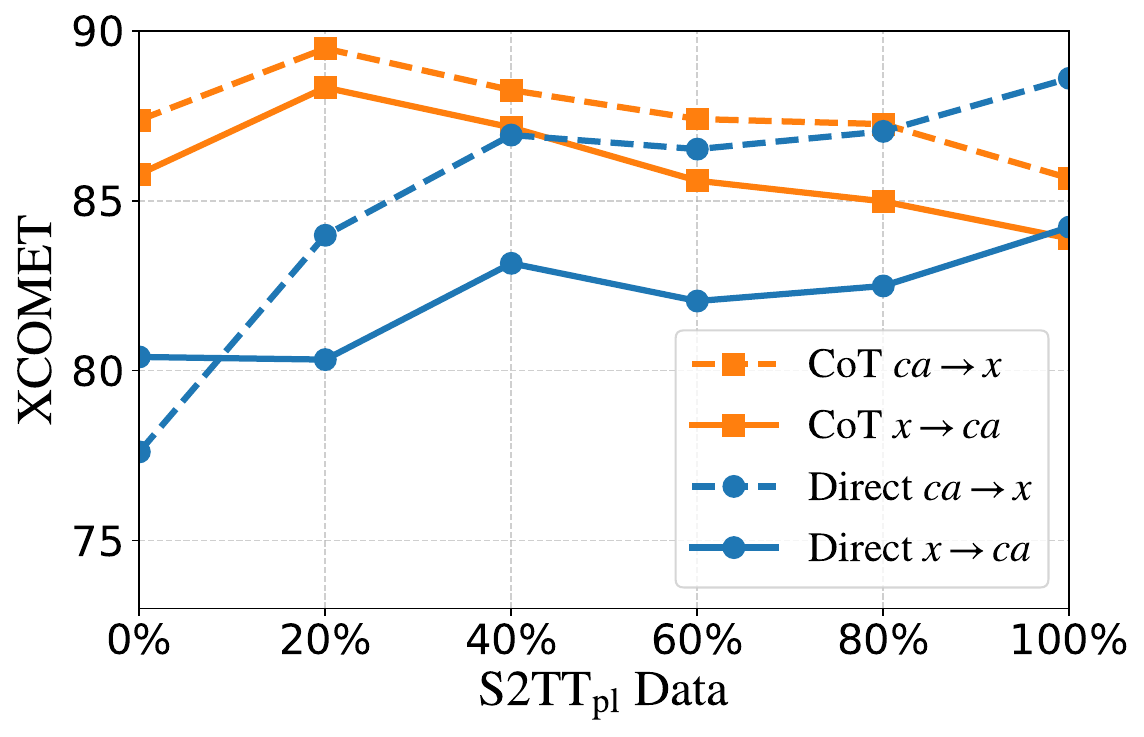}
    \caption{\textit{ca$\rightarrow$x} and \textit{x$\rightarrow$ca}}
    \label{fig:s2tt_plots_ca}
  \end{subfigure}
  \hfill
  \begin{subfigure}{0.32\textwidth}
    \centering
    \includegraphics[width=\linewidth]{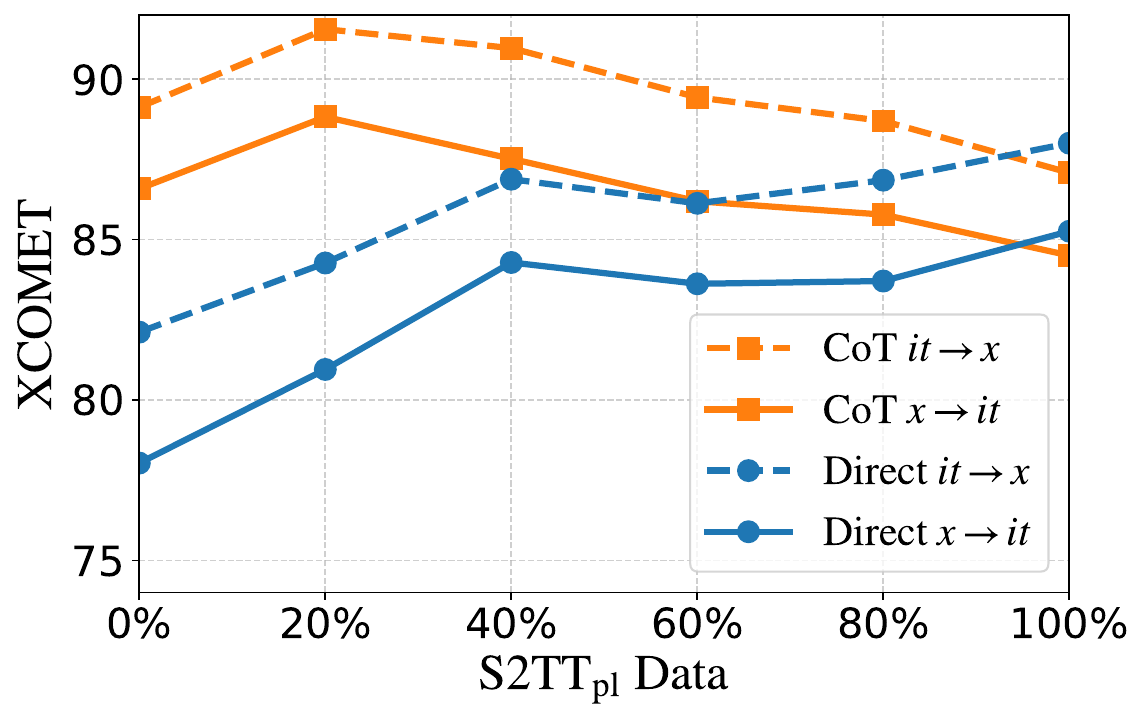}
    \caption{\textit{it$\rightarrow$x} and \textit{x$\rightarrow$it}}
    \label{fig:s2tt_plots_it}
  \end{subfigure}
      \caption{S2TT results evaluated on Fleurs test set with x\textsc{comet}. The x-axis represents the percentage of the S2TT\textsubscript{pl} data with which the different \directname\xspace and \cotname\xspace models were trained. Note that $0\%$ corresponds to \textsc{base} variants and the rest to \textsc{aug\textsubscript{XX}}.}
  \label{fig:s2tt_plots}
  \vspace{-4pt}
\end{figure*}

\noindent\textbf{\cotpaugmented lags behind \cotaugmented}\;~Figures~\ref{fig:s2tt_plot} and \ref{fig:asr_plot} support the hypothesis from \S\nobreak\ref{subsec:experminents} that training the transcription step may negatively impact the model, as \cotpaugmented consistently underperforms \cotaugmented in ASR and S2TT. Nonetheless, \cotaugmented also leads to some degradation. A possible explanation is that, by not training the transcription step, all S2TT\textsubscript{pl} data is treated as T2TT with speech as context, effectively reducing the proportion of the ASR task.

\noindent\textbf{Results are consistent across languages}\;~To assess whether these results hold across languages, Figure~\ref{fig:s2tt_plots} shows S2TT performance for \textit{en}, \textit{ca}, and \textit{it}. We selected these languages because \textit{en} benefits from the the majority of ASR data and typically achieves strong translation performance due to the abundance of English-centric datasets. The other two are extremes in S2TT\textsubscript{pl} data quantity (see Table~\ref{tab:data-stats}): \textit{ca} has the most and \textit{it} the least.

Figure~\ref{fig:s2tt_plots_en} shows that \textit{en} follows a similar trend to Figure~\ref{fig:s2tt_plot}, though performance on \textit{en$\rightarrow$x} remains nearly constant. This is likely due to the large amount of ASR data in \textit{en}, which, similarly to~\cite{wang20ia_interspeech}, allows \directbaseline (at 0\% S2TT\textsubscript{pl} data) to perform close to \cotbaseline. Both \textit{ca} and \textit{it} exhibit the same overall trend, confirming that the results are consistent across languages. Notably, for \textit{ca$\rightarrow$x}, \directaugmentedhundred nearly matches the peak of \cotaugmentedtwenty. This supports the hypothesis that, with sufficient S2TT\textsubscript{pl} data, \directname\xspace can reach \cotname\xspace performance.

\section{Conclusions and Future Directions}
\label{sec:conclusions}

In this work, we systematically compare \cotname\xspace and \directname\xspace prompting strategies for S2TT using pseudo-labeled data, which has become the standard approach for scaling this task. Our experiments show that \cotname\xspace struggles to improve with more data, regardless of whether the transcription step is trained or not. In contrast, \directname\xspace prompting demonstrates a consistent scaling trend: although its absolute performance has not yet surpassed \cotname, it steadily improves as more pseudo-labeled data is added, indicating strong potential for larger scale trainings.

These findings position \directname\xspace prompting as a promising strategy for future research. Beyond its favorable scaling behavior, \directname\xspace also offers practical advantages: it requires roughly half the computational resources of \cotname\xspace and is simpler to implement. Moreover, the observed trends raise important questions about its limits: it remains to be validated whether \directname\xspace could not only match but eventually surpass \cotname\xspace with an even larger scale of data, clarifying its potential as a primary strategy for S2TT training.

Finally, beyond the current setup, a particularly relevant direction is paralinguistic-aware translation. E2E S2TT systems are not constrained by the transcription bottleneck, which gives them the potential to produce richer translations that reflect linguistic cues uniquely present in speech, such as prosody. However, because current training largely relies on ASR and T2TT datasets, especially in \cotname\xspace prompting, this capacity is unlikely to emerge naturally. Progress may therefore require training with direct translations annotated for speech-specific cues, and it would be valuable to investigate whether \directname\xspace prompting can better exploit the full richness of the speech signal compared to \cotname\xspace.


\ninept
\section{Acknowledgements}

This work is funded by the Ministerio para la Transformación Digital y de la Función Pública and Plan de Recuperación, Transformación y Resiliencia -- Funded by EU -- NextGenerationEU within the framework of the project Modelos del Lenguaje. FC and CEB acknowledge their AI4S fellowship within the “Generación D” initiative by Red.es, Ministerio para la Transformación Digital y de la Función Pública, for talent attraction (C005/24-ED CV1), funded by NextGenerationEU through PRTR.

\bibliographystyle{IEEEbib_mod}
\bibliography{refs}

@inproceedings{du-etal-2025-making,
    title = "Making {LLM}s Better Many-to-Many Speech-to-Text Translators with Curriculum Learning",
    author = "Du, Yexing  and
      Pan, Youcheng  and
      Ma, Ziyang  and
      Yang, Bo  and
      Yang, Yifan  and
      Deng, Keqi  and
      Chen, Xie  and
      Xiang, Yang  and
      Liu, Ming  and
      Qin, Bing",
    editor = "Che, Wanxiang  and
      Nabende, Joyce  and
      Shutova, Ekaterina  and
      Pilehvar, Mohammad Taher",
    booktitle = "Proceedings of the 63rd Annual Meeting of the Association for Computational Linguistics (Volume 1: Long Papers)",
    month = jul,
    year = "2025",
    address = "Vienna, Austria",
    publisher = "Association for Computational Linguistics",
    url = "https://aclanthology.org/2025.acl-long.610/",
    doi = "10.18653/v1/2025.acl-long.610",
    pages = "12466--12478",
    ISBN = "979-8-89176-251-0"
}

@inproceedings{zhang_speechgpt_2023,
	address = {Singapore},
	title = {{SpeechGPT}: {Empowering} {Large} {Language} {Models} with {Intrinsic} {Cross}-{Modal} {Conversational} {Abilities}},
	shorttitle = {{SpeechGPT}},
	url = {https://aclanthology.org/2023.findings-emnlp.1055/},
	doi = {10.18653/v1/2023.findings-emnlp.1055},
	urldate = {2025-02-06},
	booktitle = {Findings of the {Association} for {Computational} {Linguistics}: {EMNLP} 2023},
	publisher = {Association for Computational Linguistics},
	author = {Zhang, Dong and Li, Shimin and Zhang, Xin and Zhan, Jun and Wang, Pengyu and Zhou, Yaqian and Qiu, Xipeng},
	editor = {Bouamor, Houda and Pino, Juan and Bali, Kalika},
	month = dec,
	year = {2023},
	pages = {15757--15773},
}

@misc{rubenstein_audiopalm_2023,
	title = {{AudioPaLM}: {A} {Large} {Language} {Model} {That} {Can} {Speak} and {Listen}},
	shorttitle = {{AudioPaLM}},
	url = {http://arxiv.org/abs/2306.12925},
	language = {en},
	urldate = {2024-11-04},
	publisher = {arXiv},
	author = {Rubenstein, Paul K. and Asawaroengchai, Chulayuth and Nguyen, Duc Dung and Bapna, Ankur and Borsos, Zalán and Quitry, Félix de Chaumont and Chen, Peter and Badawy, Dalia El and Han, Wei and Kharitonov, Eugene and Muckenhirn, Hannah and Padfield, Dirk and Qin, James and Rozenberg, Danny and Sainath, Tara and Schalkwyk, Johan and Sharifi, Matt and Ramanovich, Michelle Tadmor and Tagliasacchi, Marco and Tudor, Alexandru and Velimirović, Mihajlo and Vincent, Damien and Yu, Jiahui and Wang, Yongqiang and Zayats, Vicky and Zeghidour, Neil and Zhang, Yu and Zhang, Zhishuai and Zilka, Lukas and Frank, Christian},
	month = jun,
	year = {2023},
	note = {arXiv:2306.12925 [cs]},
}

@article{hsu_hubert_2021,
	title = {{HuBERT}: {Self}-{Supervised} {Speech} {Representation} {Learning} by {Masked} {Prediction} of {Hidden} {Units}},
	volume = {29},
	issn = {2329-9290},
	shorttitle = {{HuBERT}},
	url = {https://doi.org/10.1109/TASLP.2021.3122291},
	doi = {10.1109/TASLP.2021.3122291},
	urldate = {2025-02-19},
	journal = {IEEE/ACM Trans. Audio, Speech and Lang. Proc.},
	author = {Hsu, Wei-Ning and Bolte, Benjamin and Tsai, Yao-Hung Hubert and Lakhotia, Kushal and Salakhutdinov, Ruslan and Mohamed, Abdelrahman},
	month = oct,
	year = {2021},
	pages = {3451--3460},
}

@INPROCEEDINGS{wu-et-al-ondecoderonly,
  author={Wu, Jian and Gaur, Yashesh and Chen, Zhuo and Zhou, Long and Zhu, Yimeng and Wang, Tianrui and Li, Jinyu and Liu, Shujie and Ren, Bo and Liu, Linquan and Wu, Yu},
  booktitle={2023 IEEE Automatic Speech Recognition and Understanding Workshop (ASRU)}, 
  title={On Decoder-Only Architecture For Speech-to-Text and Large Language Model Integration}, 
  year={2023},
  volume={},
  number={},
  pages={1-8},
  doi={10.1109/ASRU57964.2023.10389705}}

@inproceedings{chen-etal-2024-llast,
    title = "{LL}a{ST}: Improved End-to-end Speech Translation System Leveraged by Large Language Models",
    author = "Chen, Xi  and
      Zhang, Songyang  and
      Bai, Qibing  and
      Chen, Kai  and
      Nakamura, Satoshi",
    editor = "Ku, Lun-Wei  and
      Martins, Andre  and
      Srikumar, Vivek",
    booktitle = "Findings of the Association for Computational Linguistics: ACL 2024",
    month = aug,
    year = "2024",
    address = "Bangkok, Thailand",
    publisher = "Association for Computational Linguistics",
    url = "https://aclanthology.org/2024.findings-acl.416/",
    doi = "10.18653/v1/2024.findings-acl.416",
    pages = "6976--6987"
}

@inproceedings{hu2025chain,
  author={Hu, Ke and Chen, Zhehuai and Yang, Chao-Han Huck and Żelasko, Piotr and Hrinchuk, Oleksii and Lavrukhin, Vitaly and Balam, Jagadeesh and Ginsburg, Boris},
  booktitle={ICASSP 2025 - 2025 IEEE International Conference on Acoustics, Speech and Signal Processing (ICASSP)}, 
  title={Chain-of-Thought Prompting for Speech Translation}, 
  year={2025},
  volume={},
  number={},
  pages={1-5},
  doi={10.1109/ICASSP49660.2025.10890560}
}

@misc{huang_speech_2023,
	title = {Speech {Translation} with {Large} {Language} {Models}: {An} {Industrial} {Practice}},
	shorttitle = {Speech {Translation} with {Large} {Language} {Models}},
	url = {http://arxiv.org/abs/2312.13585},
	doi = {10.48550/arXiv.2312.13585},
	urldate = {2025-01-29},
	publisher = {arXiv},
	author = {Huang, Zhichao and Ye, Rong and Ko, Tom and Dong, Qianqian and Cheng, Shanbo and Wang, Mingxuan and Li, Hang},
	month = dec,
	year = {2023},
	note = {arXiv:2312.13585 [cs]},
}

@misc{du_cot-st_2024,
	title = {{CoT}-{ST}: {Enhancing} {LLM}-based {Speech} {Translation} with {Multimodal} {Chain}-of-{Thought}},
	shorttitle = {{CoT}-{ST}},
	url = {http://arxiv.org/abs/2409.19510},
	doi = {10.48550/arXiv.2409.19510},
	urldate = {2025-01-29},
	publisher = {arXiv},
	author = {Du, Yexing and Ma, Ziyang and Yang, Yifan and Deng, Keqi and Chen, Xie and Yang, Bo and Xiang, Yang and Liu, Ming and Qin, Bing},
	month = sep,
	year = {2024},
	note = {arXiv:2409.19510 [cs]},
}

@inproceedings{wang20ia_interspeech,
  title     = {Improving Cross-Lingual Transfer Learning for End-to-End Speech Recognition with Speech Translation},
  author    = {Changhan Wang and Juan Pino and Jiatao Gu},
  year      = {2020},
  booktitle = {Interspeech 2020},
  pages     = {4731--4735},
  doi       = {10.21437/Interspeech.2020-2955},
  issn      = {2958-1796},
}

@inproceedings{pino-etal-2019-harnessing,
    title = "Harnessing Indirect Training Data for End-to-End Automatic Speech Translation: Tricks of the Trade",
    author = "Pino, Juan  and
      Puzon, Liezl  and
      Gu, Jiatao  and
      Ma, Xutai  and
      McCarthy, Arya D.  and
      Gopinath, Deepak",
    editor = {Niehues, Jan  and
      Cattoni, Rolando  and
      St{\"u}ker, Sebastian  and
      Negri, Matteo  and
      Turchi, Marco  and
      Ha, Thanh-Le  and
      Salesky, Elizabeth  and
      Sanabria, Ramon  and
      Barrault, Loic  and
      Specia, Lucia  and
      Federico, Marcello},
    booktitle = "Proceedings of the 16th International Conference on Spoken Language Translation",
    month = nov # " 2-3",
    year = "2019",
    address = "Hong Kong",
    publisher = "Association for Computational Linguistics",
    url = "https://aclanthology.org/2019.iwslt-1.18/"
}

@article{hannun2014deep,
  title={Deep Speech: Scaling up end-to-end speech recognition},
  author={Hannun, Awni Y and Case, Carl and Casper, Jared and Catanzaro, Bryan and Diamos, Greg and Elsen, Erich and Prenger, Ryan and Satheesh, Sanjeev and Sengupta, Shubho and Coates, Adam and Ng, Andrew Y},
  journal={arXiv preprint arXiv:1412.5567},
  year={2014}
}

@inproceedings{ardila_common_2020,
	address = {Marseille, France},
	title = {Common {Voice}: {A} {Massively}-{Multilingual} {Speech} {Corpus}},
	isbn = {979-10-95546-34-4},
	shorttitle = {Common {Voice}},
	url = {https://aclanthology.org/2020.lrec-1.520/},
	language = {eng},
	urldate = {2025-02-13},
	booktitle = {Proceedings of the {Twelfth} {Language} {Resources} and {Evaluation} {Conference}},
	publisher = {European Language Resources Association},
	author = {Ardila, Rosana and Branson, Megan and Davis, Kelly and Kohler, Michael and Meyer, Josh and Henretty, Michael and Morais, Reuben and Saunders, Lindsay and Tyers, Francis and Weber, Gregor},
	editor = {Calzolari, Nicoletta and Béchet, Frédéric and Blache, Philippe and Choukri, Khalid and Cieri, Christopher and Declerck, Thierry and Goggi, Sara and Isahara, Hitoshi and Maegaard, Bente and Mariani, Joseph and Mazo, Hélène and Moreno, Asuncion and Odijk, Jan and Piperidis, Stelios},
	month = may,
	year = {2020},
	pages = {4218--4222},
}

@article{pratap2020MLSAL,
  title={MLS: A Large-Scale Multilingual Dataset for Speech Research},
  author={Vineel Pratap and Qiantong Xu and Anuroop Sriram and Gabriel Synnaeve and Ronan Collobert},
  journal={ArXiv},
  year={2020},
  volume={abs/2012.03411}
}

@InProceedings{TIEDEMANN12.463,
  author = {Jörg Tiedemann},
  title = {Parallel Data, Tools and Interfaces in OPUS},
  booktitle = {Proceedings of the Eight International Conference on Language Resources and Evaluation (LREC'12)},
  year = {2012},
  month = {may},
  date = {23-25},
  address = {Istanbul, Turkey},
  editor = {Nicoletta Calzolari (Conference Chair) and Khalid Choukri and Thierry Declerck and Mehmet Ugur Dogan and Bente Maegaard and Joseph Mariani and Jan Odijk and Stelios Piperidis},
  publisher = {European Language Resources Association (ELRA)},
  isbn = {978-2-9517408-7-7},
  language = {english}
 }

@inproceedings{
    kargaran2023glotlid,
    title={{G}lot{LID}: Language Identification for Low-Resource Languages},
    author={Kargaran, Amir Hossein and Imani, Ayyoob and Yvon, Fran{\c{c}}ois and Sch{\"u}tze, Hinrich},
    booktitle={The 2023 Conference on Empirical Methods in Natural Language Processing},
    year={2023},
    url={https://openreview.net/forum?id=dl4e3EBz5j}
}

@inproceedings{dale-costa-jussa-2024-blaser,
    title = "{BLASER} 2.0: a metric for evaluation and quality estimation of massively multilingual speech and text translation",
    author = "Dale, David  and
      Costa-juss{\`a}, Marta R.",
    editor = "Al-Onaizan, Yaser  and
      Bansal, Mohit  and
      Chen, Yun-Nung",
    booktitle = "Findings of the Association for Computational Linguistics: EMNLP 2024",
    month = nov,
    year = "2024",
    address = "Miami, Florida, USA",
    publisher = "Association for Computational Linguistics",
    url = "https://aclanthology.org/2024.findings-emnlp.943/",
    doi = "10.18653/v1/2024.findings-emnlp.943",
    pages = "16075--16085"
}

@INPROCEEDINGS{jairsan2020a,
  author={J. {Iranzo-Sánchez} and J. A. {Silvestre-Cerdà} and J. {Jorge} and N. {Roselló} and A. {Giménez} and A. {Sanchis} and J. {Civera} and A. {Juan}},
  booktitle={ICASSP 2020 - 2020 IEEE International Conference on Acoustics, Speech and Signal Processing (ICASSP)}, 
  title={Europarl-ST: A Multilingual Corpus for Speech Translation of Parliamentary Debates}, 
  year={2020},
  pages={8229-8233},
}

@inproceedings{wang_covost_2021,
	title = {{CoVoST} 2 and {Massively} {Multilingual} {Speech} {Translation}},
	url = {https://www.isca-archive.org/interspeech_2021/wang21s_interspeech.html},
	doi = {10.21437/Interspeech.2021-2027},
	urldate = {2025-02-13},
	booktitle = {Proc. {Interspeech} 2021},
	author = {Wang, Changhan and Wu, Anne and Gu, Jiatao and Pino, Juan},
	year = {2021},
	pages = {2247--2251},
}

@inproceedings{conneau_fleurs_2023,
	title = {{FLEURS}: {Few}-{Shot} {Learning} {Evaluation} of {Universal} {Representations} of {Speech}},
	shorttitle = {{FLEURS}},
	url = {https://ieeexplore.ieee.org/document/10023141},
	doi = {10.1109/SLT54892.2023.10023141},
	urldate = {2025-02-13},
	booktitle = {2022 {IEEE} {Spoken} {Language} {Technology} {Workshop} ({SLT})},
	author = {Conneau, Alexis and Ma, Min and Khanuja, Simran and Zhang, Yu and Axelrod, Vera and Dalmia, Siddharth and Riesa, Jason and Rivera, Clara and Bapna, Ankur},
	month = jan,
	year = {2023},
	pages = {798--805},
}

@inproceedings{papineni_bleu_2002,
	address = {Philadelphia, Pennsylvania, USA},
	title = {Bleu: a {Method} for {Automatic} {Evaluation} of {Machine} {Translation}},
	shorttitle = {Bleu},
	url = {https://aclanthology.org/P02-1040/},
	doi = {10.3115/1073083.1073135},
	urldate = {2025-02-20},
	booktitle = {Proceedings of the 40th {Annual} {Meeting} of the {ACL}},
	publisher = {Association for Computational Linguistics},
	author = {Papineni, Kishore and Roukos, Salim and Ward, Todd and Zhu, Wei-Jing},
	editor = {Isabelle, Pierre and Charniak, Eugene and Lin, Dekang},
	month = jul,
	year = {2002},
	pages = {311--318},
}

@article{guerreiro-etal-2024-xcomet,
    title = "xcomet: Transparent Machine Translation Evaluation through Fine-grained Error Detection",
    author = "Guerreiro, Nuno M.  and
      Rei, Ricardo  and
      Stigt, Daan van  and
      Coheur, Luisa  and
      Colombo, Pierre  and
      Martins, Andr{\'e} F. T.",
    journal = "Transactions of the Association for Computational Linguistics",
    volume = "12",
    year = "2024",
    address = "Cambridge, MA",
    publisher = "MIT Press",
    url = "https://aclanthology.org/2024.tacl-1.54/",
    doi = "10.1162/tacl_a_00683",
    pages = "979--995"
}

@inproceedings{hassid_textually_2023,
	title = {Textually {Pretrained} {Speech} {Language} {Models}},
	url = {https://openreview.net/forum?id=UlHueVjAKr&noteId=6gD38fPDxO},
	language = {en},
	urldate = {2025-02-06},
	booktitle = {Thirty-seventh {Conference} on {Neural} {Information} {Processing} {Systems}},
	author = {Hassid, Michael and Remez, Tal and Nguyen, Tu Anh and Gat, Itai and Conneau, Alexis and Kreuk, Felix and Copet, Jade and Défossez, Alexandre and Synnaeve, Gabriel and Dupoux, Emmanuel and Schwartz, Roy and Adi, Yossi},
	month = nov,
	year = {2023},
}

@inproceedings{post_call_2018,
	address = {Brussels, Belgium},
	title = {A {Call} for {Clarity} in {Reporting} {BLEU} {Scores}},
	url = {https://aclanthology.org/W18-6319/},
	doi = {10.18653/v1/W18-6319},
	urldate = {2025-02-20},
	booktitle = {Proceedings of the {Third} {Conference} on {Machine} {Translation}: {Research} {Papers}},
	publisher = {Association for Computational Linguistics},
	author = {Post, Matt},
	editor = {Bojar, Ondřej and Chatterjee, Rajen and Federmann, Christian and Fishel, Mark and Graham, Yvette and Haddow, Barry and Huck, Matthias and Yepes, Antonio Jimeno and Koehn, Philipp and Monz, Christof and Negri, Matteo and Névéol, Aurélie and Neves, Mariana and Post, Matt and Specia, Lucia and Turchi, Marco and Verspoor, Karin},
	month = oct,
	year = {2018},
	pages = {186--191},
}

@misc{Duquenne:2023:sonar_arxiv,
  author = {Paul-Ambroise Duquenne and Holger Schwenk and Benoit Sagot},
  title = {{SONAR:} Sentence-Level Multimodal and Language-Agnostic Representations},
  publisher = {arXiv},
  year = {2023},
  url = {https://arxiv.org/abs/2308.11466},
}

\end{document}